\documentclass[10pt,conference]{IEEEtran}
\IEEEoverridecommandlockouts
\usepackage{cite}
\usepackage{amsmath,amssymb,amsfonts}
\usepackage{algorithmic}
\usepackage{graphicx}
\usepackage{textcomp}
\usepackage{xcolor}

\usepackage{xcolor}

\usepackage{svg}
\usepackage{amsmath}  
\usepackage{soul, xcolor}   
\usepackage{listings}
\usepackage{tabularx}
\usepackage{amssymb}

\usepackage[flushleft]{threeparttable} 
\usepackage{booktabs,caption}
\usepackage{graphicx}
\newcolumntype{C}[1]{>{\centering}p{#1}}
\usepackage{microtype}

\usepackage[T1]{fontenc}
\usepackage[utf8]{inputenc}
\usepackage{mathtools}
\usepackage{blkarray, bigstrut}
\usepackage{gauss}
\usepackage{array}
    
\begin{document}

\title{\huge \textit{"I think this is the most disruptive technology"} \newline
Exploring Sentiments of ChatGPT Early Adopters \newline using Twitter Data}

\author{\IEEEauthorblockN{Mubin Ul Haque, Isuru Dharmadasa, Zarrin Tasnim Sworna, Roshan Namal Rajapakse, Hussain Ahmad*}
\thanks{All the authors contributed equally to this paper.\newline *Corresponding author: hussain.ahmad@adelaide.edu.au}

\IEEEauthorblockA{\{mubinul.haque, isuru.mahaganiarachchige, zarrintasnim.sworna, roshan.rajapakse, hussain.ahmad\}@adelaide.edu.au\\School of Computer Science, University of Adelaide, Australia}

}


\maketitle
\thispagestyle{plain}
\pagestyle{plain}

\begin{abstract}
Large language models have recently attracted significant attention due to their impressive performance on a variety of tasks. ChatGPT developed by OpenAI is one such implementation of a large, pre-trained language model that has gained immense popularity among early adopters, where certain users go to the extent of characterizing it as a \textit{disruptive technology} in many domains. Understanding such early adopters' sentiments is important because it can provide insights into the potential success or failure of the technology, as well as its strengths and weaknesses. In this paper, we conduct a mixed-method study using 10,732 tweets from early ChatGPT users. We first use topic modelling to identify the main topics and then perform an in-depth qualitative sentiment analysis of each topic. Our results show that the majority of the early adopters have expressed overwhelmingly positive sentiments related to topics such as \textit{Disruptions to software development}, \textit{Entertainment and exercising creativity}. Only a limited percentage of users expressed concerns about issues such as the potential for misuse of ChatGPT, especially regarding topics such as \textit{Impact on educational aspects}. We discuss these findings by providing specific examples for each topic and then detail implications related to addressing these concerns for both researchers and users.
\end{abstract}

\begin{IEEEkeywords}
ChatGPT, Generative Pretrained Transformer, Early adopters, Twitter, Sentiment Analysis, Topic Modeling
\end{IEEEkeywords}
\section{Introduction}

ChatGPT is an artificial intelligence (AI) chatbot that understands and generates natural human language with remarkable sophistication, sensitivity, and usability \cite{chatgpt_human_replace}. ChatGPT is an application of the latest version of GPT-3 (Generative Pretrained Transformer 3), a state-of-the-art language processing AI model developed by the \textit{OpenAI}\footnote{https://openai.com/} foundation, that enables it to generate human-like text. Unlike traditional chatbots, ChatGPT remembers what the user said earlier in the conversation for follow-up questions, rejects inappropriate requests, and challenges incorrect responses \cite{chatgpt_explain}. Moreover, ChatGPT provides answers, solutions, and descriptions to complex questions, including potential ways to solve layout problems, write code, and answer optimization queries \cite{chatgpt_code}. Given the advantages of ChatGPT over traditional chatbots, ChatGPT has attracted more than 1 million users in just one week after it was launched, leaving behind other popular online platforms such as Netflix, Facebook, and Instagram in terms of adoption rates \cite{chatgpt_tweet}. In addition, there is a rising number of commentators who predict that ChatGPT will replace Google in the near future \cite{friedman_2022}. Some early adopters of ChatGPT believe that it will eventually obsolete several professions related to content creation, such as programmers, professors, playwrights, and journalists \cite{chatgpt_human_replace}. For example, it has been demonstrated that ChatGPT is capable of producing high-quality responses to a variety of challenges, including solving coding challenges and generating accurate responses to exam queries \cite{chatgpt_writing}. 

As described above, ChatGPT offers several benefits to a wide range of users. However, being a new technology, identifying early adopter sentiments is of high importance due to several reasons. Firstly, early adopters are usually the most enthusiastic and influential users of a product, and their opinions and sentiments can help to shape the broader perception of new technology. This information can provide critical insights into the potential success or failure of the product.

Secondly, early adopters are often the first to encounter any issues or problems with new technology, and their feedback can help to identify and fix these issues before they become widespread. Therefore, exploring the early adopter sentiments, particularly for a disruptive technology such as ChatGPT, would increase the chances of success of the tool in the market.

\begin{figure*} [t]
  \centering
  \includegraphics[width=\textwidth]{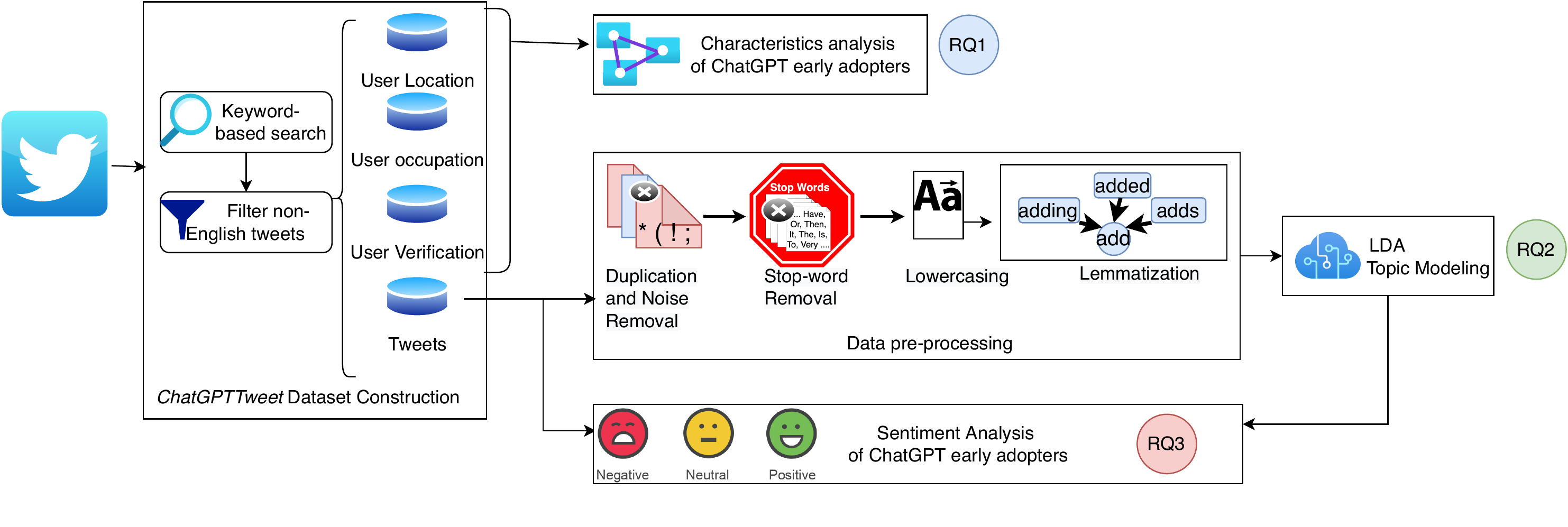}
  \caption{Overview of our research methodology}
  \label{fig:method} 
\end{figure*}

To investigate the sentiments of ChatGPT early adopters, we analyze data from Twitter, which allows users to read and share 140-character messages called ``tweets". Unlike other social networking platforms (e.g., Facebook and Instagram), unregistered users can access and read tweets \cite{anber2016literature}. Moreover, Twitter is a famous and large social networking micro-blogging site \cite{sahayak2015sentiment}. With the burgeoning popularity of Twitter, researchers and practitioners are increasingly using Twitter data to get untapped information from potential customers \cite{bian2016mining}. For example, Twitter has been explored for investigating the public perception of \textit{"Internet of Things"} \cite{bian2016mining}, COVID-19 symptoms emergence \cite{guo2020mining}, drug-related adverse events detection \cite{bian2012towards}, work emotion and stress analysis \cite{wang2016twitter}, mining public health data \cite{paul2011you}, and influenza epidemics detection \cite{aramaki2011twitter}. Therefore, in this study, we decided to use Twitter as a data source to analyze the sentiments of early adopters of ChatGPT.

In this paper, we conduct a mixed-method study on 10,732 tweets from early adopters. We first use topic modelling to identify the main topics of discussion. Next, we perform manual sentiment analysis to qualitatively analyze a selected set of tweets from each topic. 

In summary, our paper makes the following contributions.


\begin{itemize}

\item It provides an overarching analysis of topics discussed by early adopters of ChatGPT through their tweets. We describe each discussed topic in detail with specific tweet examples. Moreover, we show the total number of tweets against each topic to demonstrate its significance compared to other topics.
    
\item It presents a high-level investigation of the sentiments of early ChatGPT adopters for each identified topic. The users' sentiments are categorized based on positive, negative, and neutral. This analysis shows the users' perception of each topic.
   
\item We describe potential future research directions and areas where ethical and societal implications are present regarding ChatGPT use.
    
\end{itemize}


The rest of the paper is organized as follows. Section \ref{sec:relatedwork} presents the related work of this study. While Section \ref{sec:method} reports the research methodology, Section \ref{sec:result} describes the results. Section \ref{Implications} presents the implications of our study for researchers and users. Section \ref{sec:threats} describes the threats to the validity of this study. Lastly, the conclusion of this study is presented in Section \ref{sec: conclusion}.

\section{Related Work} \label{sec:relatedwork}

As ChatGPT technology is quite new, we have not found substantial research directly related to ChatGPT. However, there exist literature on the GPT family of text-generating AIs (e.g., GPT-2 and GPT-3). Also, we identified sufficient literature on Twitter data mining to investigate users' sentiments. Therefore, in the following, we report some closely related studies to our study.

Bian et al. \cite{bian2016mining} analyzed Twitter data to understand the public sentiments about \textit{Internet of Things} (IoT). They performed sentiment analysis and topic modelling to identify prominent topics and public attitudes toward each topic. As a result, they found users are more interested in business and technology compared to other domains of IoT, and have favourable sentiments toward IoT. Similarly, Trivedi et al. \cite{trivedi2022mining} introduced a \textit{Robustly Optimized BERT Pre-training Approach (RoBERTa)} to investigate public sentiments through their tweets on hybrid work arrangements. The \textit{RoBERTa} revealed that the majority of users have positive sentiments for the hybrid work model. Another study \cite{alhijawi2022prediction} reported Twitter temporal data mining for predicting a movie's success. The authors proposed a rating prediction model and a temporal product popularity model to forecast users' satisfaction and movie popularity among users, respectively. Similarly, Nawaz et al. \cite{nawaz2022mining} proposed a prediction model to predict the political election results of Pakistan. The authors showed 98\% accuracy and efficiency of their model in predicting the results through Twitter data as compared to alternative approaches. 

To the best of our knowledge, our work is the first study that conducted a qualitative analysis of early adopter sentiments and feedback on ChatGPT. We contribute to the literature by providing a snapshot of the early public responses to this latest technology.



\section{Research Methodology}\label{sec:method}
This section presents our research methodology, where we discuss our research questions, dataset construction, pre-processing steps, identification of discussed ChatGPT topics, and sentiment analysis on these topics. Fig. \ref{fig:method} presents the overview of our research methodology.

\begin{figure*} [t]
  \centering
  \includegraphics[width=1\textwidth]{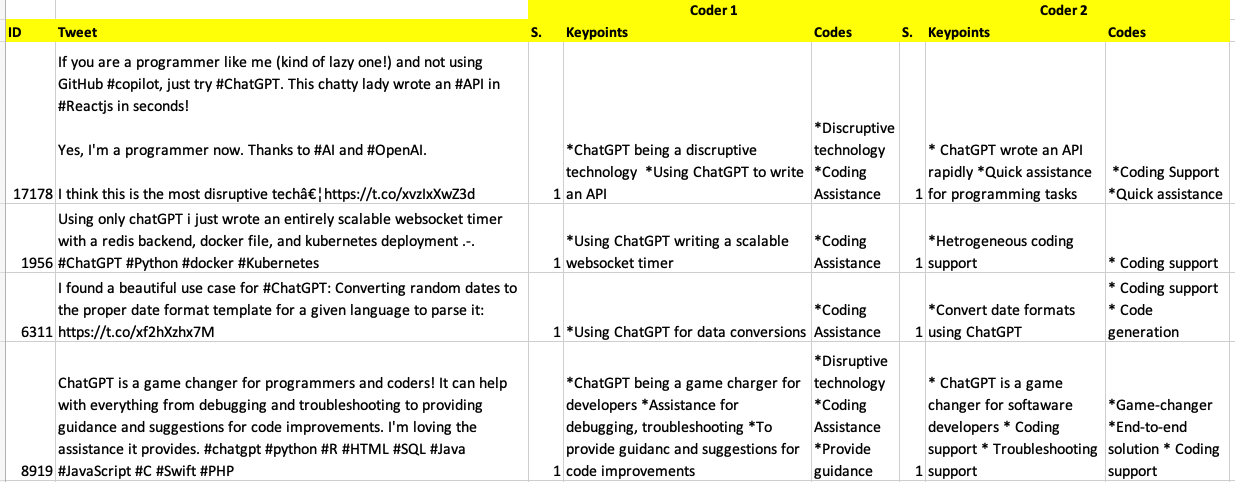}
  \caption{Qualitative analysis of tweets: Initial version of the code sheet for T1}
  \label{fig:coding} 
\end{figure*}

\subsection{Research Questions} \label{subsec:rq}
The following RQs motivated our empirical study.
\begin{itemize}
    \item \textbf{RQ1}. What are the characteristics of ChatGPT early adopters?
    \item \textbf{RQ2}. ChatGPT Topics - What are the main topics that are being discussed about ChatGPT on Twitter?
    \item \textbf{RQ3}. ChatGPT Sentiments - What are the sentiments that are being expressed about ChatGPT topics on Twitter?
\end{itemize}

\subsection{ChatGPT Tweet Dataset Construction}
To assess public sentiments on the early adoption of ChatGPT, we collected social media data, specifically from Twitter. We collected tweets from December 5, 2022 to December 7, 2022. While collecting the tweets, we only considered the tweets that included the keyword "ChatGPT". Besides, we only collected the tweets that were written in the English language. We used Python and Twitter API to extract Twitter data. In our $ChatGPTTweet$ dataset, we have 18K tweets, where we collected the text, user location, user occupation, user verification, posted date, and hashtags for each tweet. To answer RQ1, we analyzed user location, user occupation, and user verification status information.

\subsection{Data Pre-processing}
We pre-processed our $ChatGPTTweet$ dataset using the following steps:

\subsubsection{Duplication Removal} We removed retweets and duplicate tweets where a user repeated a tweet of another user. After duplicate tweet removal, our ChatGPTTweet dataset includes 10,732 tweets that we used in our analysis.

\subsubsection{Lowercasing} We lowercased the tweets that represent words in different cases (e.g., StackOverflow and Stackoverflow) to the same lower-case form (e.g., StackOverflow). 

\subsubsection{Noise Removal} We removed noises (e.g., punctuation marks) to retain only the alphanumerical data for cleaning our $ChatGPTTweet$ dataset. In the noise removal step, we also removed URLs, Emojis, and Twitter handles.

\subsubsection{Stop words Removal} 
We removed stop-words that appear frequently (e.g., this, are, and a) but do not help to distinguish one tweet from another. We removed stop-words using the NLTK \cite{nltkbook} English stop-word list. Besides, we removed the top three frequently appearing key domain specific words, such as ChatGPT, OpenAI, and AI.

\subsubsection{Lemmatization}
We performed WordNet-based lemmatization using NLTK \cite{nltkbook}. We used lemmatization to represent a word's inflected forms (e.g., getting, gets) to its dictionary-based root form (e.g., get), which is a common practice in the existing literature \cite{apiro}. 


\begin{figure*}
  \centering
  \includegraphics[trim=0 140 290 0,clip, scale = 0.6]{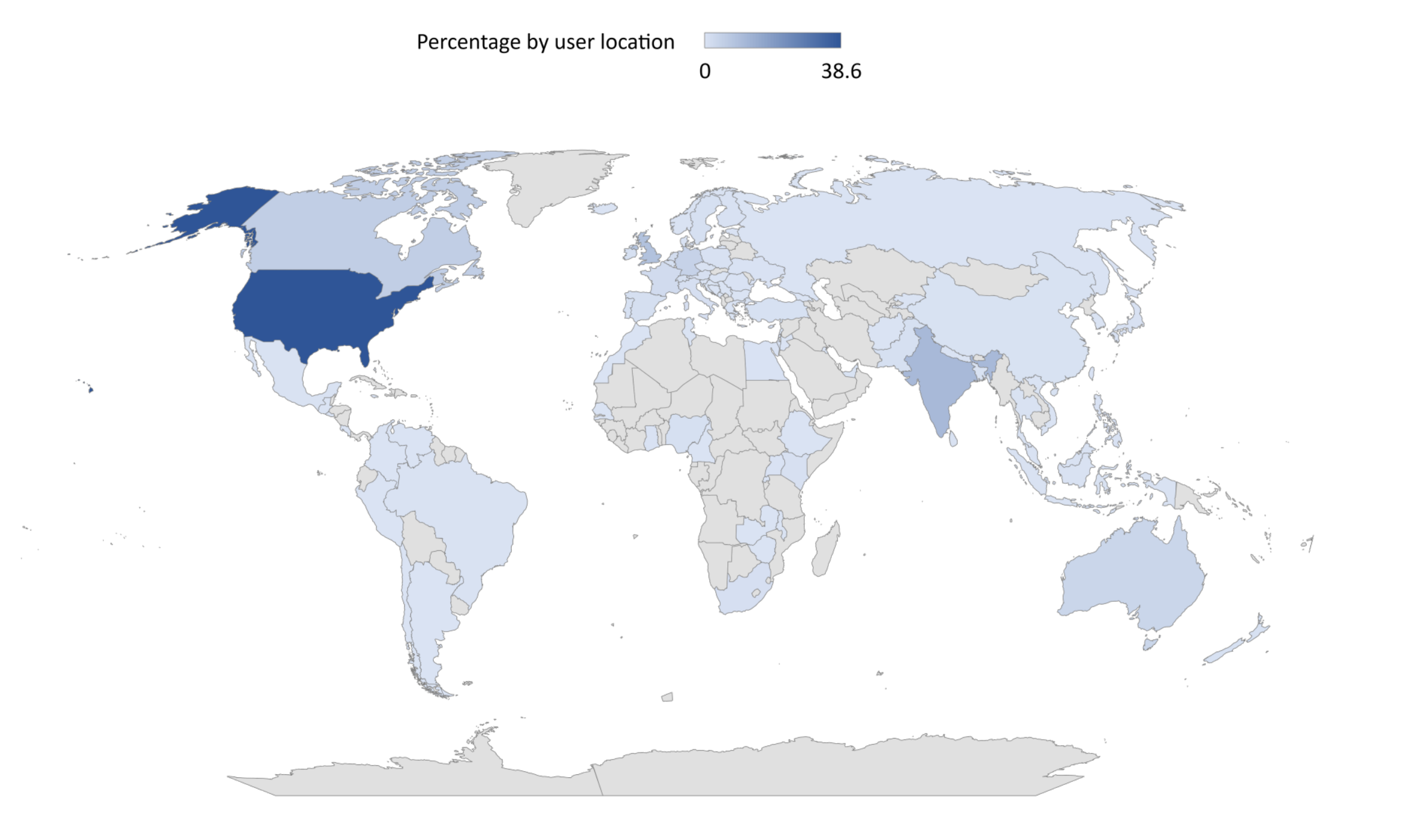}
  \caption{Geographic distribution of the early adopters of ChatGPT based on tweets}
  \label{fig:loc} 
\end{figure*}

\subsection{Identification of ChatGPT Topics} To answer RQ2, we identified a set of ChatGPT key topics using the Latent Dirichlet Allocation (LDA) modelling technique \cite{LDA}. LDA is a commonly used technique for topic modelling in the existing Software Engineering (SE) literature \cite{docker, zahedi2020mining}. LDA is used to group tweets of our $ChatGPTTweet$ dataset into a set of topics using word co-occurrence and frequency. A set of probabilities are assigned to each tweet by LDA. Here, the probabilities refer to the chances of a tweet being related to a specific topic. We used MALLET \cite{MALLET} implementation of LDA, which is commonly used in existing literature \cite{concurrency, bigdata}. 


Identification of the optimal number of topics $N$, while implementing LDA is critical as LDA may generate a huge number of narrow topics for a high value of $N$. 
In contrast, LDA may create broad generalized topics for a low value of $N$. 
Hence, we executed a broad range of experiments to analyze the coherence scores, which were achieved by varying $N$ values from 5 to 60, with steps=2 and iteration=100, 500, and 1000 using MALLET. 
The coherence score depicts the understandability of LDA topics that is relevant to human comprehensibility \cite{roder2015exploring}. 
We obtained a comparatively high coherence score by running MALLET when the number of topics ranged from 9 to 12.

%
To ensure that we select the optimal number of topics, we examined randomly selected 15 tweets from each topic for 9 $\leq$ $N$ $\leq$ 12. 
Based on our examination, we found that $N$ = 9 (i.e, the optimal number of topics is 9) serves the purpose of balancing the comprehensibility of our dataset. 
%
This approach is followed in similar studies to find the optimal number of topics \cite{zahedi2020mining}. We then ran LDA using MALLET with N=9 and generated a CSV file for each topic. Each CSV file was sorted based on the highly relevant documents of the topic from LDA topic modeling method.

\subsection{Sentiment Analysis on ChatGPT Topics} \label{sec:senti} To answer RQ3, we performed sentiment analysis for each of our identified topics on ChatGPT. For this task, we initially used Python's NLTK Library to automatically classify the tweets per topic. However, upon inspecting the results, we were not satisfied with some of the automated classifications of the library. For example, certain tweets that we identified as neutral (discussing both negative and positive aspects of ChatGPT) were classified as negative by the library. In addition, a substantial amount of tweets were accompanied by images (e.g., Screenshots). We preferred to have a look at the complete tweet with these images for an overall view of the content of the tweet. Finally, our previous work, which explored the most recent trends and emerging solutions of a highly evolving field, showed that qualitative analysis enables a more in-depth and nuanced data analysis \cite{rajapakse2022collaborative}. Therefore, we chose to do a manual qualitative analysis of the tweets. We aim to extend this work with the inclusion of a larger dataset in the future to assess overall trends with time, where we plan to use an automated analysis method.

For sentiment analysis, we manually labelled 9 datasets, where each sample dataset ($S_i$, where $i$ ranges from 1 to 9) includes 100 randomly selected tweets for a specific topic using MS excel. We labelled a tweet with -1, 0, and 1, where -1 represents negative sentiment, 1 represents positive sentiment, and 0 represents neutral sentiment. We kept the neutral sentiment label for tweets that discuss both positive and negative aspects and tweets that were ambiguous to identify a specific positive or negative sentiment. We then did open coding \cite{strauss1997grounded} for each tweet to enable us to identify the common patterns of discussion within a topic. In this task, we identified \textit{keypoints} (i.e., summarised
points) and then \textit{codes} (i.e., a phrase that further summarizes
the key point in 2 or 3 words) and recorded them in the same spreadsheets as an example snapshot is shown in Fig. \ref{fig:coding}.

All five authors were involved in the manual labelling and qualitative analysis of our data. Here, each data set per topic was labelled by two authors. All disagreements were resolved through open discussions among the annotators to reach an agreement for mitigating any possible errors. 
%


\section{Result Analysis}\label{sec:result}

We describe the results for our RQ1, RQ2, and RQ3 in Section \ref{RQ1}, \ref{RQ2}, and \ref{RQ3}, respectively.


\subsection{RQ1. What are the characteristics of ChatGPT early adopters?}
\label{RQ1}

\textbf{Motivation} In this RQ, we aim to analyze the characteristics of the early adopters of ChatGPT. We analyze the characteristics in terms of user location, occupation and account verification status. This analysis will enable us to focus on the characteristics of the early adopters to understand users from which professions are interested in ChatGPT and the demographic distribution of these users.

\textbf{Approach} We quantitatively analyze the user location, user occupation, and user verification status of our $ChatGPTTweet$ dataset to answer this RQ.

\textbf{Result}
For user location, Fig. \ref{fig:loc} demonstrates that early adopters are geographically dispersed, where the majority of the tweets originated from the North American and Asian regions, and a similar number of tweets from Europe, Australia and South America. 
In our analysis, we found USA, India, UK, Canada, and Germany are the top-5 countries in terms of expressing their opinion while adopting ChatGPT.

\begin{figure}
  \centering
  \includegraphics[width=\columnwidth]{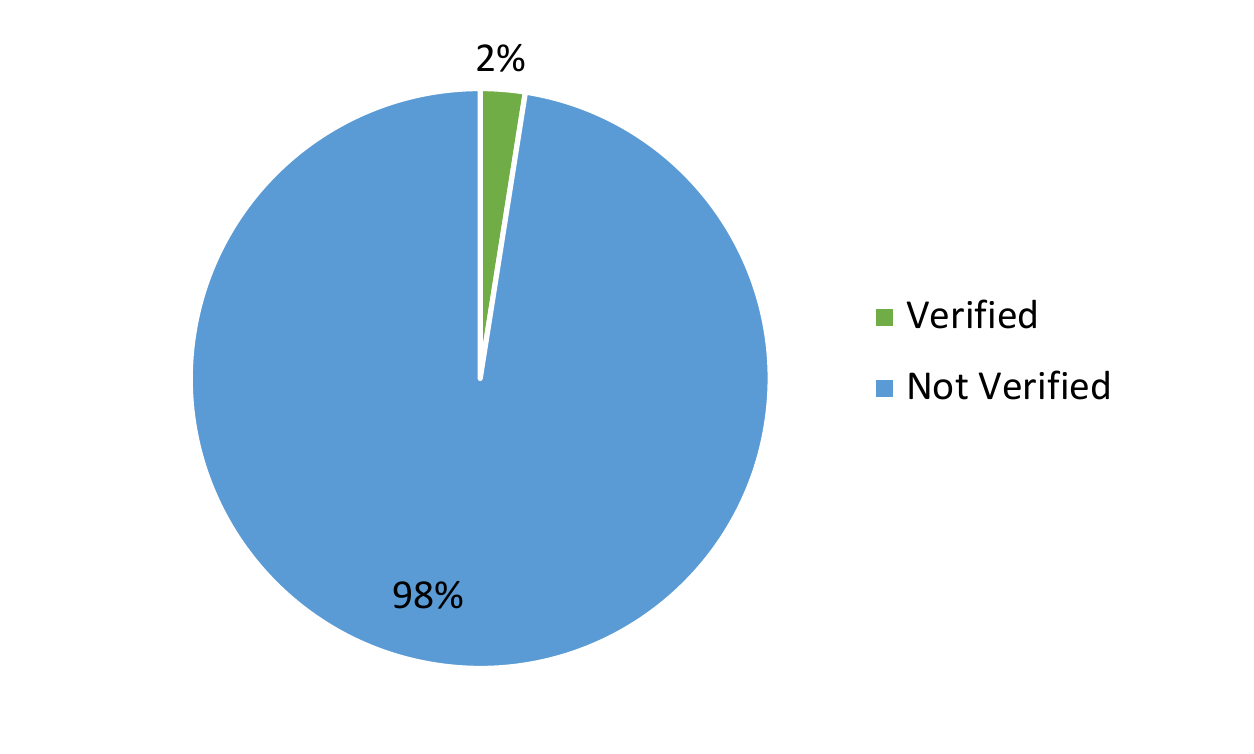}
  \caption{Distribution of the verification status of the early adopters of ChatGPT based on Twitter}
  \label{fig:verified} 
\end{figure}

Besides, our analysis identifies that only 2\% of the early adopters of ChatGPT are verified Twitter users as shown in Fig. \ref{fig:verified}. It implies that ChatGPT is not confined only to verified users; rather, it is widely adopted across all Twitter users.

\begin{figure} 
  \centering
  \includegraphics[width=\columnwidth]{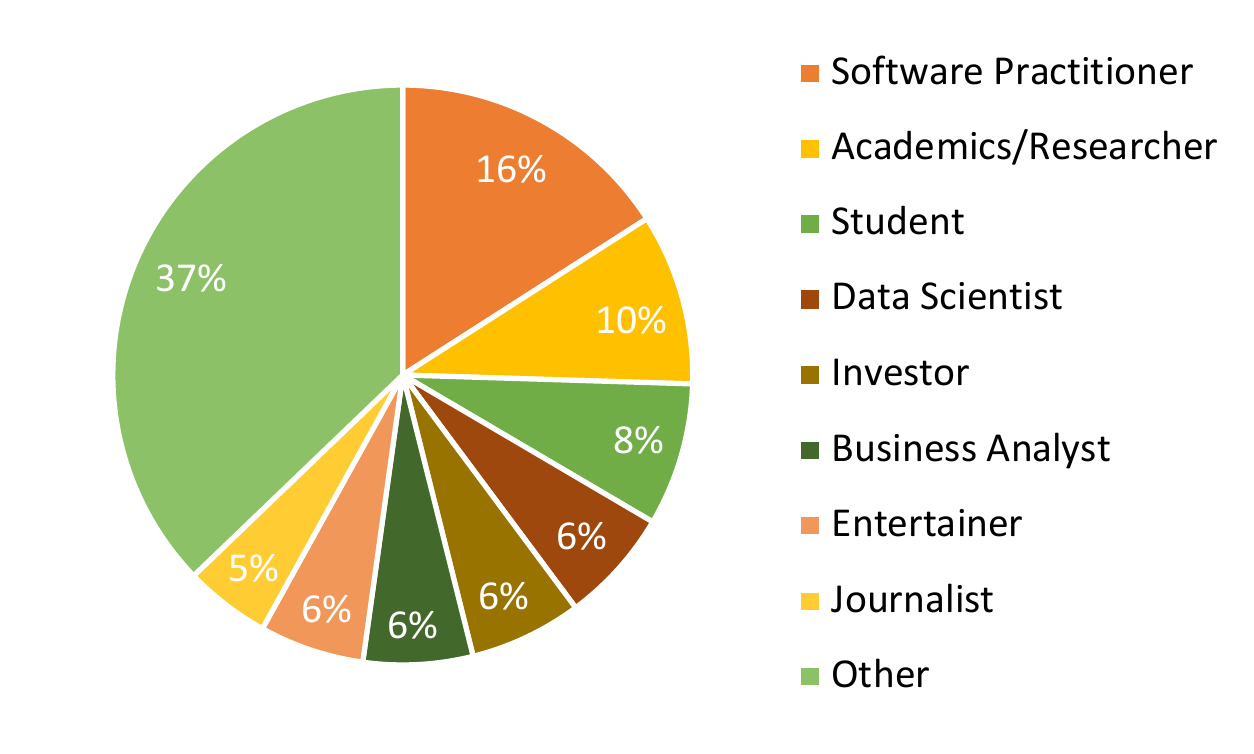}
  \caption{Occupation distribution of the early adopters of ChatGPT based on tweets}
  \label{fig:jobanalysis} 
\end{figure}

For user occupation, we identified a broad, wide, and diverse range of early adopter communities for ChatGPT. In our analysis, we also identified Twitter users with no specific user job description, which we showed as \textit{other} in our occupation distribution that is depicted in Fig. \ref{fig:jobanalysis}. The top 3 user occupations of the early adopters of ChatGPT are software practitioners, academics, and students. Other occupations include data scientists, investors, business analysts, entertainers (e.g., artists, singers, actors), and journalists, as shown in Fig. \ref{fig:jobanalysis}. It implies the popularity and interest in adopting AI-based ChatGPT across diverse occupations.

For RQ1, we identified that the early adopters of ChatGPT are located in geographically dispersed regions with a diverse and broad range of professions.

%


\subsection{RQ2. ChatGPT Topics - What are the main topics that are being discussed about ChatGPT in Twitter?}
\label{RQ2}
\textbf{Motivation.} ChatGPT has attained significant attraction over broad and diverse communities, which includes not only researchers, managers, and practitioners, but also entertainers, business analysts, and educationists. 

For instance, the registered number of users for ChatGPT has risen up to 1 Million within 5 days since its beta-release \cite{chatgpt_tweet}, whereas Facebook, Netflix, and Instagram required approximately 300, 1200, and 75 days to reach 1 Million users. 
An answer to this RQ will identify the most common and pressing ChatGPT topics that the communities have frequently encountered while using ChatGPT.
This identification will enable us to understand how diverse communities express their experience and insights over different domains. 
We assert that this understanding is invaluable to assess ChatGPT's capability, effectiveness, and usability.

\textbf{Approach.} We identified the optimal number of topics for our $ChatGPT$ dataset by using LDA as discussed in Section \ref{sec:method}. 
After identifying the optimal number of topics, we examined the top 20 keywords and randomly selected 30 tweets from each of the identified topics to select a suitable name, which provides the best representation for the group of tweets under that topic. 
All the authors were involved in the examination and discussed extensively to reach a consensus on naming the topics.
This approach for naming the topics is common and widely adopted in the literature \cite{docker,zahedi2020mining,bigdata,concurrency,wan2019discussed,abdellatif2020challenges}. 

\begin{figure} 
  \centering
  \includegraphics[width=\columnwidth]{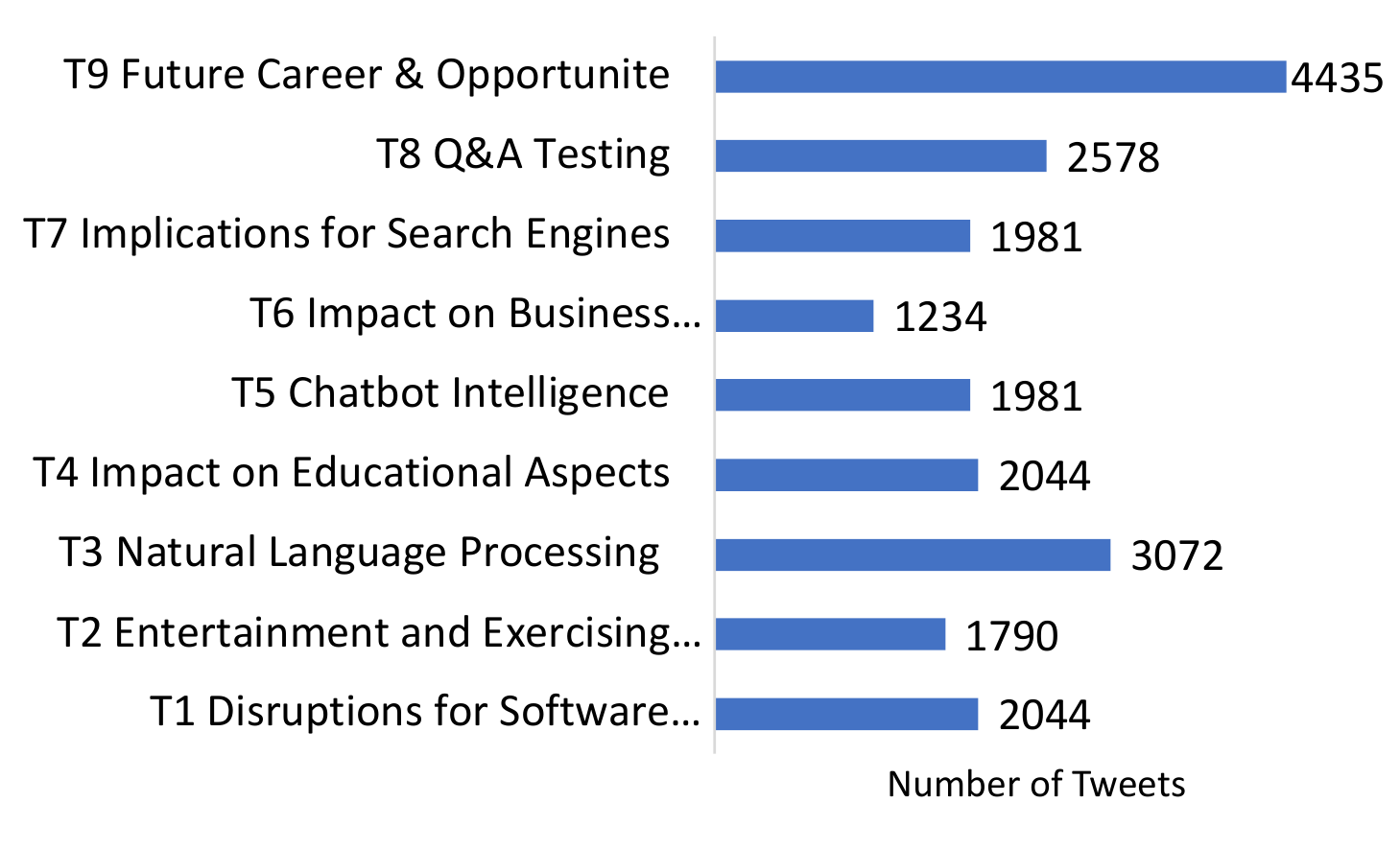}
  \caption{Distribution of topics based on the number of tweets captured by Topic Modeling}
  \label{fig:numTweet} 
\end{figure}

\textbf{Results.} We identified 9 topics as shown in Table \ref{tab:topic} that have been discussed among the early adopters of ChatGPT. Fig. \ref{fig:numTweet} depicts the distribution of these topics based on the number of tweets captured by Topic Modelling. We describe these results in detail below.
%
%

\begin{table*}[ht]
\centering
\caption{Topic Names and top 10 words (lemmatized topic words) for our $ChatGPTTweet$ topics}
\label{tab:topic}
\resizebox{\textwidth}{!}{%
\begin{tabular}{lll}
\hline
Sl & Topic Name & Topic keywords \\
\hline
T1 & { Disruptions for Software Development} & {code, write, create, program, generate, python, script, developer, error, run} \\
T2 & { Entertainment and Exercising Creativity} & { write, story, poem, love, fun, short, joke, style, funny, movie} \\
T3 & { Natural Language Processing} & { model, language, generate, data, text, prompt, human, conversation, learn, response} \\
T4 & {Impact on Educational Aspects} & { write, student, paper, essay, plan, research, education, school, assignment, teach, homework} \\
T5 & { Chatbot Intelligence} & { chatbot, intelligence, artificialintelligence, machinelearning, artificial, user, million, robot, security, app} \\
T6 & { Impact on Business development} & { time, startup, business, company, service, true, idea, control, market, customer} \\
T7 & { Implications for Search Engines} & { google, search, answer, engine, replace, result, source, StackOverflow, query, internet, information, avlable} \\
T8 & { Q\&A Testing} & { question, answer, wrong, test, response, correct, amp, pretty, simple, solve}\\
T9 & { Future Careers \& Opportunities} & { tool, future, time, people, technology, potential, job, world, change, learn}\\
\hline
\end{tabular}
}
\end{table*}

\subsubsection{Disruptions to software development} \label{SE}
A key area discussed in the extracted dataset is the potential disruptions ChatGPT will cause to current software development practices. For example, users discussed many examples of how ChatGPT can be used to generate code, assist with debugging, and even perform tasks like summarizing and translating code. 

\subsubsection{Entertainment and exercising creativity}
Twitter users were widely using ChatGPT for entertainment purposes, generating poems, jokes or other humorous write-ups. To generate entertaining outputs using a ChatGPT, users need to provide the model with some initial text to work with, such as a prompt or seed text. A popular use case in this area was efforts to combine characteristics of different entities (e.g., movie or TV characters, popular personalities and concepts) in one amusing write-up.
 


\subsubsection{Natural Language Processing}

The promising Natural Language Processing (NLP) capabilities of ChatGPT enable it to generate understandable natural human language that facilitates ChatGPT users in perceiving, comprehending, and projecting generated text. Moreover, the ChatGPT NLP capability significantly enhances its usability, utility, and efficiency. Therefore, we have observed that early adopters have meticulously discussed the NLP aspect of ChatGPT in the extracted dataset. For example, users have discussed the amalgamation of AI and NLP in ChatGPT technology in their tweets. This enables ChatGPT to generate and understand human-like texts as an outcome.

\subsubsection{Impact on Educational Aspects} The early adopters of ChatGPT are considering ChatGPT as one of the technologies to change the traditional way of education. Users are discussing how ChatGPT can be used for different purposes in the education domain. For example, users discussed the use of ChatGPT for early childhood learning, developing syllabi, scientific literature review, and crisis management learning, such as safety plans for thoughts of suicide.

\subsubsection{Chatbot Intelligence}

Twitter data reveals that the intelligence of ChatGPT has become a prominent discussion point among early adopters. The AI capability of ChatGPT enables it to understand and respond to users' queries in a meaningful manner. With the help of AI, ChatGPT can entertain quite complex and challenging problems that traditional chatbots fail to understand accurately. For example, ChatGPT can write and debug complex codes, solve large-scale optimization problems, and provide accurate responses to complicated queries.

\subsubsection{Impacts on Business Development} \label{Business}

The application of ChatGPT in business analysis is being widely discussed on Twitter. Users are discussing the use of ChatGPT for developing startup pitches, generating business use cases, and creating business plans. Users are also asking ChatGPT to provide financial advice.

\subsubsection{Implications for Search Engines}
One of the widely discussed topics is using ChatGPT as a search engine to query and retrieve a wide range of information. Exceeding the existing search engine (e.g., Google Search, Bing) functionalities, ChatGPT has provided users with a novel experience of presenting information conveniently by selecting the most appropriate information and explaining it in simple terms.   

\subsubsection{Q\&A Testing} Q\&A Testing is the third largest topic in our $ChatGPTTweet$ dataset. 
In this topic, adopters typically use ChatGPT to learn, compare, and verify answers for different academic subjects (e.g., physics, mathematics, chemistry), or/and conceptual subjects (e.g., philosophy, religion), among others.
%
%
We identified adopters who also ask open-ended and analytical questions to ChatGPT in order to understand the capability of ChatGPT. Furthermore, we observed several questions on complex technical and emerging subjects as well.
%
%
%
\subsubsection{Future Careers \& Opportunities} This topic is the largest discussed topic for ChatGPT, where early adopters share their insights and ideas on how this advent AI technology can influence future career opportunities.
Early adopters also discussed what we require in terms of skills, knowledge, values, and behaviour to cope-up with this advanced technology.
Besides, we identified early adopters' significant insight on how the industry and research organizations can collaborate to stay up-to-date on the latest developments in AI and ensure that sustainable growth and progress are achieved and remain relevant to the social, emotional, and cognitive aspects of humans. 

\subsection{RQ3. ChatGPT Sentiments - What are the sentiments that are being expressed about ChatGPT topics on Twitter?}
\label{RQ3}
\textbf{Motivation.} Earlier research efforts \cite{bian2016mining, trivedi2022mining} in analyzing Twitter sentiments (i.e., positive, negative, or neutral opinion) asserted that the sentiment analysis is extremely helpful in determining the public or user perception towards a product, service, or technology.
Sentiment analysis is also significant since it can impact the longevity of a product, service, and technology \cite{pak2010twitter,gokulakrishnan2012opinion,wakade2012text}.
As a highly trending chatbot technology, user communities expressed their sentiments while using ChatGPT on Twitter. 
Answers to this RQ will help the ChatGPT decision makers, where they need to act promptly, as Twitter sentiments typically provide emotion-rich information.
%

\begin{figure*} 
  \centering
  \includegraphics[width=0.8\textwidth]{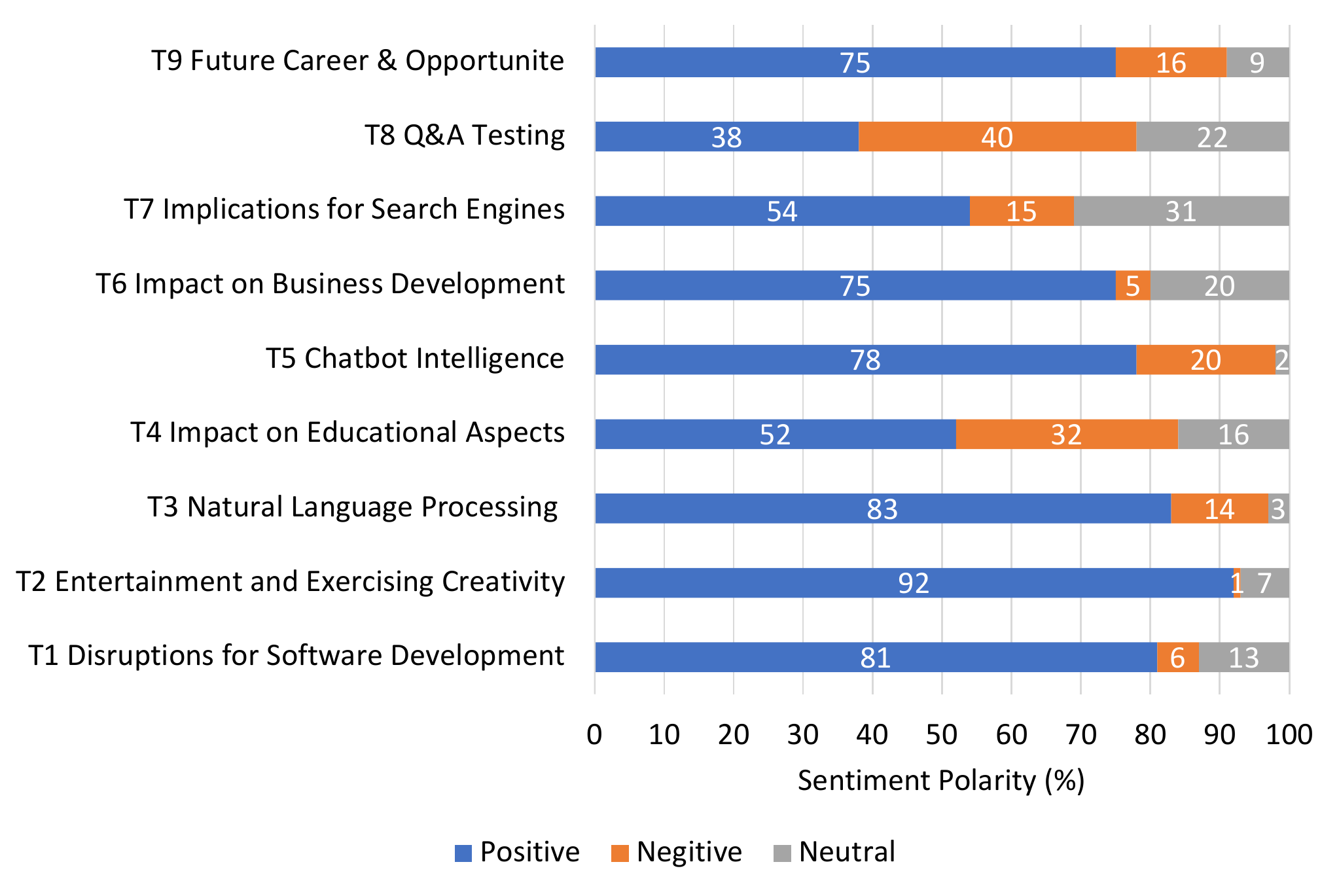}
  \caption{Results of the qualitative sentiment analysis per topic}
  \label{SentimentAnalysis} 
\end{figure*}

\textbf{Approach.} We followed the approach described in Section \ref{sec:senti}.

\textbf{Results.} Fig. \ref{SentimentAnalysis} depicts the summary of the percentage values (Positive, Negative and Neutral) for each topic. Below, we describe the results in detail for each topic.

\subsubsection{Disruptions to software development}

The sentiment analysis returned 81\% positive for this topic. Early adopters were particularly impressed by ChatGPT's abilities to assist in coding tasks. We captured many tweets of users giving specific examples of how ChatGPT assisted them with their software development activities.

\begin{quote}
 \textit{If you are a programmer like me (kind of lazy one!) and not using GitHub \#copilot, just try \#ChatGPT. This chatty lady wrote an API in \#Reactjs in seconds! 
Yes, I'm a programmer now. Thanks to \#AI and \#OpenAI.
\newline
I think this is the most disruptive tech [..].}  
\end{quote}

Another popular use case with positive sentiments was the assistance this technology provides with regard to debugging and error handling. Users expressed their satisfaction with how ChatGPT provided them with specific assistance with sorting out their coding-related troubleshooting as compared with the other available services.

\begin{quote}
 \textit{ChatGPT is a game changer for programmers and coders! It can help with everything from debugging and troubleshooting to providing guidance and suggestions for code improvements. I'm loving the assistance it provides. \#chatgpt \#python \#R \#HTML \#SQL \#Java \#JavaScript \#C \#Swift \#PHP}
\end{quote}

Only a very less amount of users (i.e., 6) expressed concerns about the disruptions ChatGPT will cause to the software development practices. Some users noted that users need to be mindful of how they use ChatGPT for specific development assistance.

\begin{quote} \textit{As many have said, \#chatGPT runs the risk of spitting out plausible-sounding but wrong answers in some cases.
It's especially bad for generating code that uses external libraries that change over time, as one might expect.
Simple logic is good tho}\end{quote}

We categorized this tweet as neutral as it points out both positive and negative aspects of ChatGPT use. Overall, users are particularly positive that this technology has the potential to save developers time and effort in their day-to-day tasks drastically.

\subsubsection{Entertainment and exercising creativity}

Naturally, the sentiment analysis for this topic turned out to be overwhelmingly positive (i.e., 92\%). Twitter users were widely using ChatGPT for entertainment purposes, generating poems, jokes or other humorous write-ups. To generate entertaining outputs using a ChatGPT, users need to provide the model with some initial text to work with, such as a prompt or seed text. A popular use in this area was efforts to combine characteristics of different entities (e.g., movie or TV characters, popular personalities and concepts) in one amusing write-up.

\begin{quote}
 \textit{I asked \#ChatGPT “In the style of a Shakespearean play, write a scene from the sitcom Friends, in which Chandler accuses Joey of stealing his banana which he had been saving to eat for his lunch.”}
 \end{quote}

Users also attempted to exercise their creativity by generating interesting ideas, such as short stories or poems.

\begin{quote}
 \textit{\#ChatGPT write a short story about a dog named Baxter who has died and is in heaven enjoying cheeseburgers and beaches, who would like to solve the mystery of the stolen Christmas dinner ham, in the style of Arthur Conan Doyle.}
\end{quote}

We observed that a screenshot of the ChatGPT query output is usually accompanied with these tweets. By analyzing the selected tweets for sentiment analysis on this topic on Twitter itself, we recognised that the ChatGPT-generated output was very positively received as they had a high engagement. This might have contributed to the drastic rise in the popularity of this technology, as evidenced by the large initial number of subscribers.
 

\subsubsection{Natural Language Processing.}

We observed diverse sentiments (i.e., positive, negative, and neutral) of early adopters of ChatGPT technology in our Twitter dataset. While most tweets (i.e., 83\%) imply satisfaction with ChatGPT, some tweets (14\%) have expressed their concerns. Only 3\% of the tweets reported a neutral viewpoint. For example, early adopters seem happy with the amalgam of NLP and AI in ChatGPT technology.

\begin{quote}
 \textit{I'm really impressed with the natural language processing capabilities of ChatGPT's AI-powered chatbot. It's a great example of the power of \#NLP \#AI technology. \#chatgpt}  
\end{quote}

Early adopters were also surprised by the realistic human-like text generation of ChatGPT.

\begin{quote}
\textit{\#ChatGPT is a prototype dialogue-based \#AI chatbot capable of understanding natural human language and generating impressively detailed human-like written text.}  
\end{quote}

The use of human AI trainers and supervised fine-tuning improves the ChatGPT text generation and perception. However, some users have shown their concerns about the quality of the generated text. They argue that ChatGPT provides misinformation due to the lack of critical-thinking skills, nuance, and ethical decision-making ability of ChatGPT.

\begin{quote}
\textit{I can always spot generated text - it lacks the depth and complexity of human thought}  
\end{quote}

In summary, though early adopters raised some negative sentiments, the majority of ChatGPT users showed positive responses. We believe ChatGPT is a promising stepping stone to the development of an absolute human-like chatbot. 

\subsubsection{Chatbot Intelligence.}

In the extracted dataset, we have identified both the positive and negative sentiments of early adopters regarding the intelligence capability of ChatGPT. As per our quantitative analysis, 78\% tweets showed a favourable sentiment toward ChatGPT intelligence, 20\% tweets raised harmful impacts of AI for ChatGPT, and 2\% tweets had a neutral opinion. This shows that the majority of early adopters are in favour of ChatGPT intelligence capability.

\begin{quote}
\textit{[...] ChatGPT could be a big help and see the amazing wonders of AI (Artificial Intelligence) in the future. @OpenAI
I typed "Create React App" and it provided me a clear to read instructions on creating the application. 
\#chatGPT}
\end{quote}

On the other hand, early adopters have also raised their negative sentiments about AI involvement in ChatGPT. For example, users were concerned about the ominous impacts of AI-based ChatGPT on society such as an increase in terrorism, hacking, and unemployment.  

\begin{quote}
\textit{ChatGPT artificial intelligence developed by OpenAI explains how to bomb and theft}
\end{quote}

In conclusion, ChatGPT's intelligence capability provides both positive and negative impacts on society. However, we recommend an effective policy/regulation development for the usage of ChatGPT. This mitigates the negative impacts of ChatGPT intelligence and hence improves ChatGPT acceptability among its users in the future.

\subsubsection{Impact on Educational Aspects} Unlike the other topics (e.g., software development) where ChatGPT is accepted overwhelmingly, the adoption of ChatGPT for educational purposes raised both positive and negative perceptions among the users. Our analysis identified 52\% positive, 32\% negative, and 16\% neutral views on the use of ChatGPT for educational purposes. 

Users are accepting ChatGPT for diverse application areas in the education domain. For example, users discussed many examples of how ChatGPT can be used for grading papers, assessing students' learning, and preparing syllabi. ChatGPT is also considered as a good personal teacher for a student. Users are considering ChatGPT as one of the many technologies that will revolutionise the traditional way of teaching and assessment.

\begin{quote}
\textit{\#ChatGPT will change education as we know it. 
I am hopeful. Perhaps this is what will springboard age-old assessment practices to ones that authentically assess student learning}
\end{quote}

On the contrary, the use of ChatGPT by students for writing essays, preparing assignments, and home-works was raised as a common public concern on Twitter. Users are concerned about how the use of ChatGPT for preparing assignments by students can hinder their learning process. Other concerns include plagiarism detection for the students' assignments and shallow answers to ChatGPT in response to research questions. 

\begin{quote}
    \textit{With the rise of the amazing \#ChatGPT, I am sure many students will use it to write essays confidently yet they learn nothing. Also plagiarism software cannot detect this, even this may not be considered as plagiarism at all}
\end{quote}

In summary, though there are some negative perceptions regarding the adoption ChatGPT for preparing assignments by students, many users are arguing against this concern as they believe that ChatGPT is exceptionally good at identifying text produced by AI, which enables it to mitigate students' plagiarism. 

\begin{quote}
\textit{ChatGPT is amazing, but it will be neither an effective way for students to plagiarize nor replace humanistic education.}
\end{quote}

\subsubsection{Impact on Business Development}
The adoption of ChatGPT in business is being discussed positively with a ray of hope for future possibilities. Our analysis identified 75\% positive, 5\% negative, and 20\% neutral views on the use of ChatGPT in business. 

Users are amazed at the response of ChatGPT to build an elevator pitch for a new idea to investors which can significantly support the start-up companies. For example, startup people are hopeful to imagine ChatGPT as a technical co-founder. ChatGPT is also expected to strongly support business decision-making.

\begin{quote}
\textit{With ChatGPT and AI-assisted coding, tech businesses will become smaller and flatter. ChatGPT will help good coders become massively more productive. AND it will help management at all levels make better decisions.}
\end{quote}

Users are also discussing the use of ChatGPT to create a business plan for app development, write trading strategies, and generate business-specific use cases. Besides, users of ChatGPT are discussing diverse potential application areas of ChatGPT from a business perspective. For example, the use of ChatGPT to provide technical customer support and financial advice. 

\begin{quote}
\textit{Imagine using ChatGPT as a technical support specialist at Apple - personalized and accurate responses to customer inquiries in seconds! This is just one example of how ChatGPT can improve customer service for any company.}
\end{quote}

Our analysis identified that ChatGPT is positively accepted to have a promising future in the business domain with its positive support in all business aspects from business plan development to customer service support.

\subsubsection{Implications for Search Engines}
Our analysis identified 54\% positive, 15\% negative and 31\% neutral sentiments on this topic. It implies that most of the tweets present either a positive or neutral sentiment over negative sentiments. For example, some users have mentioned ChatGPT as a \#googlekiller, indicating that ChatGPT poses a significant threat to existing search engines. This indicates that the users tend to consider ChatGPT as a replacement for the current search engines.

\begin{quote}
 \textit{Just relaized I have been using google to access \#ChatGPT the apparent \#googlekiller.}  
\end{quote}

Furthermore, some ChatGPT users have stated that ChatGPT performs better in search speed and accuracy than the current search engines and knowledge-sharing platforms, respectively. This implies the potential future risks not only for search engine services but also for other knowledge-sharing related platforms such as Stack Overflow, Quora and Wikipedia. 

\begin{quote}
 \textit{On information side \#ChatGPT is like:
- Google but more exact
- Quora but faster
[...].}  
\end{quote}

We came across several instances of negative sentiments concerning ChatGPT's ability to provide accurate information as it is leveraging the internet as its primary data source. Furthermore, its also been remarked that some users are yet to understand these limitations of ChatGPT. Thus, users are encouraged to be critical and verify ChatGPT results from other sources before using them in their intended applications. 

\begin{quote}
 \textit{It has become evident that you cannot blindly trust \#ChatGPT; data sourced from the internet is not completely "verified" or accurate. Most people are able to recognize this fact for ChatGPT, but not for the internet as a whole. Be critical of the information you find online.}  
\end{quote}

Interestingly, several solutions (e.g., Google Chrome extension) have already emerged to address the above-identified concerns. These plugins can retrieve and present search results from ChatGPT and existing search engines, such as Google, in a single view, facilitating the user to make better-informed decisions. Therefore, we foresee ChatGPT significantly revolutionizing and enhancing the user experience through improved user interaction models in future search engines and knowledge-sharing platforms.

\subsubsection{Q\&A Testing} Our analysis identified 38\% positive, 40\% neutral, and 22\% negative sentiments while adopting ChatGPT for testing its capability in terms of  questions and answering.
The positive sentiments for \textit{Q\&A Testing} topics are evolved due to the quality, human-friendly interaction, fast responses, and the provision of reasons for the generated answers. 
For instance, we observed positive sentiments for providing users with quality answers, which enables us to get the expected value from ChatGPT. 
\begin{quote}
    \textit{I recently used \#Chatgpt to help me better understand conditional types in English, and I was impressed by the quality of the exercises it provided.
    I found that by giving it more precise and well-structured questions, I was able to get even more value out of it.}
\end{quote}
Besides, adopters are enthusiastic about the way ChatGPT interacts with the users while answering their questions. 
\begin{quote}
    \textit{This is really spot on. I played 20 questions with \#ChatGPT yesterday. First, it guessing  correctly who I and my daughter was thinking about (Rapunzel). And then us guessing that it was thinking about Billie Eilish. Really felt like an interaction with a new  kind of entity.}
\end{quote}
Another cause for the positive sentiments for this topic is the capability of ChatGPT for providing the reasons behind the answer. 
\begin{quote}
    \textit{Nurturing curiosity in children is the best thing parents, grandparents, teachers, caretakers, etc. could do. when the eventual 1st, 2nd...n "why" question hits you, might as well use \#ChatGPT instead of the catastrophic "because I told you so".}
\end{quote}
The negative sentiments for \textit{Q\&A Testing} topics are evolved due to the wrong, incorrect, and sometimes invalid answers for their questions. 
\begin{quote}
    \textit{"\#OpenAI \#ChatGPT I used this AI to find my chem[istry] answers but got the wrong answers.  for example, Answer has to be in gms, and it's giving me in moles...}
\end{quote}
Besides, we identified some adopters have justified their sentiments in this topic by mentioning that Stack Overflow has banned to post answers generated by ChatGPT.
\begin{quote}
    \textit{Stack Overflow has temporarily banned responses generated by OpenAI's \#ChatGPT AI, citing ``a high rate" of incorrect answers. ChatGPT, can answer questions about coding problems but often produces ``plausible-sounding but incorrect or nonsensical answers,".}
\end{quote}
One significant concern raised by the early adopters of ChatGPT in \textit{Q\&A} topic is the confidence shown for wrong or incorrect answers, which may hamper the ChatGPT's wider adoptability.
\begin{quote}
    \textit{The primary problem is the answers which \#ChatGPT produces have a high rate of being incorrect...The scary part was just how confidently incorrect it was.}
\end{quote}

\subsubsection{Future Careers \& Opportunities} 
Our analysis identified 75\% positive, 9\% neutral, and 16\% negative sentiments while early adopters were expressing their opinions regarding future careers and opportunities.
The positive sentiments for \textit{Future Careers \& Opportunities} topics are evolved due to the fast and effective solution provided by the ChatGPT, which can be a crucial factor for successful careers. 
\begin{quote}
    \textit{AI coding and decision support systems will enable very small teams to create massively valuable products, services, and companies.
}
\end{quote}
Besides, we observed positive sentiments for ChatGPT's innovation, learning, and adaptation to make the job tasks convenient and easy.
\begin{quote}
    \textit{AI is the future! With its ability to learn and adapt, it's no surprise that more and more companies are turning to this technology to improve their operations and better serve their customers. Exciting times ahead! \#AI \#innovation \#ChatGPT \#OpenAI}
\end{quote}
Besides, adopters are enthusiastic about the customization feature of ChatGPT to provide personalized messages, services, or answers, which adopters believed to help them to be more engaged in future with their customers. 
\begin{quote}
    \textit{Imagine using \#ChatGPT as a technical support specialist at [...] - personalized and accurate responses to customer inquiries in seconds! This is just one example of how ChatGPT can improve customer service for any company.}
\end{quote}
%
The negative sentiments for \textit{Future Careers \& Opportunities} topics are evolved due to fear among people of losing their jobs to ChatGPT. 
\begin{quote}
    \textit{\#ChatGPT ready to replace Product Managers...}
\end{quote}
Besides, we identified some adopters who have expressed to lose their software programmer jobs as ChatGPT can perform a magnitude of software development activities within a very short time, which may take a long time for a human software programmer.
\begin{quote}
    \textit{More importantly, GPT will create a huge turmoil in the IT industry. Many coding jobs will be taken over by this AI, as it is able to do a multitude of coding jobs within seconds (which takes humans many weeks to do). This is a watershed moment in tech history.}
\end{quote}
%
%

\section{Implications} \label{Implications}

Considering the increasing popularity of ChatGPT technology, we conducted this study to identify and analyze topics with regard to the sentiments of ChatGPT early adopters. The topic modelling and sentiment analysis results provide an idea of how ChatGPT technology is being perceived among early users, and it also indicates the potential for ChatGPT's acceptance in the future. In this section, we discuss implications for users and researchers based on our results.

\subsection{Implications for Users}

Our study identifies several benefits for users of ChatGPT. For example, as described in \ref{SE}, ChatGPT has the potential to change traditional ways of software development, which shows that software developers can utilize ChatGPT while creating software and ensuring an effective software development process. Similarly, as described in \ref{Business}, ChatGPT can help business developers create their feasibility reports and business cases to enable smooth operations and other business development-related tasks. For example, a user might query business ideas for a specific context, and the model could generate a list of possible ideas which can be used as a starting point for brainstorming.

Our study has recognized several pitfalls that users must consider and weigh when integrating ChatGPT into their workflows and applications. For example, since  ChatGPT results are not verified or fact-checked by any established authority, users must not solely depend on ChatGPT results to perform any critical task. Therefore, users must critically analyze ChatGPT results and consult other verified data sources to make sensible decisions. Similarly, users must be cautious about sharing their personal and confidential data with ChatGPT, as adversarial attacks on ChatGPT language models can lead to possible data breaches, leaving ChatGPT users vulnerable. It is also essential to highlight the responsible usage of ChatGPT from the users' end, as integrating ChatGPT's creations and compositions into commercial and academic works can lead to legal and ethical concerns.   

\subsection{Implications for Researchers}

Our study provides potential future research directions to researchers for exploring concerns related to the ethical use of ChatGPT and other practical implications. For example, users raised concerns about this technology being misused in writing or completing educational activities such as essays or assignments, which would hamper students' learning. This issue was captured in the \textit{Impacts on Educational Aspects} topic. Further, the topics \textit{Q\&A Testing} captured concerns related to misinformation or inaccurate outputs being generated from ChatGPT. 
These issues indicate that researchers should explore means of mitigating these ethical and practical implications. For example, effective protocols for ChatGPT's usage can be developed to ensure its ethical usage across different domains.

\section{Threats to validity}\label{sec:threats}
Firstly, our study focuses on tweets from Twitter as a representative of the sentiments of early adopters of ChatGPT. Twitter was the 9th most visited website globally in 2021 \cite{twitter_stat}, which is an enriched knowledge base for analyzing human sentiment. However, future research can focus on other resources, such as Stack Overflow and blogs, to further generalize our findings.

Secondly, manual topic-wise sentiment analysis can be subject to human judgment bias. To mitigate this threat to validity, each sample dataset of a specific topic was labelled by two researchers, and any disagreements were resolved through discussion.

Thirdly, while threat modelling is useful in handling a large amount of data, its usage introduces some threats. For example, identifying the optimal value for a number of topics $N$ is a potential threat. To mitigate it, we followed the commonly used method of experimentation with a broad range of values for $N$ \cite{stevens2012exploring, docker}.

\section{Conclusion}\label{sec: conclusion}

In conclusion, our study of early adopters' sentiments about ChatGPT revealed overwhelming excitement and limited concerns about this application of a large language model. The majority of users were impressed by the performance of ChatGPT and the potential of large language models to assist with tasks related to several domains (e.g., Software development, Business initiatives and analysis, NLP). However, there are also important ethical implications that need to be considered in ChatGPT use and further development. For example, some users were concerned about the negative effect it would have on the education industry activities such as take-home assignments and essay writing for students. Overall, our study provides valuable insights into the sentiments of early adopters of ChatGPT and highlights the need for continued research and dialogue to develop best practices for the responsible use of large language models.


\bibliographystyle{IEEEtran}
\bibliography{conference_101719}

\end{document}